\documentclass[letterpaper,journal]{IEEEtran}
\usepackage{xcolor} 
\usepackage{amsmath,amsfonts}
\usepackage{array}
\usepackage[caption=false,font=normalsize,labelfont=sf,textfont=sf]{subfig}
\usepackage{textcomp}
\usepackage{stfloats}
\usepackage{url}
\usepackage{verbatim}
\usepackage{graphicx}
\usepackage{algpseudocode}
\usepackage{booktabs}
\usepackage{multirow}
\usepackage[colorlinks=true,linkcolor=blue,citecolor=blue,urlcolor=blue]{hyperref}
\usepackage{cite}

\usepackage{cuted}
\usepackage{capt-of}

\makeatletter
\renewcommand{\@cite}[2]{%
  \textcolor{blue}{[%
  #1\if@tempswa , #2\fi%
  ]}%
}
\makeatother

\def\BibTeX{{\rm B\kern-.05em{\sc i\kern-.025em b}\kern-.08em
    T\kern-.1667em\lower.7ex\hbox{E}\kern-.125emX}}
\usepackage{balance}

\begin{document}

\title{\LARGE \bf
ExoGS: A 4D Real-to-Sim-to-Real Framework\\for Scalable Manipulation Data Collection
}

\author{Yiming Wang, Ruogu Zhang, Minyang Li, Hao Shi,
Junbo Wang, Deyi Li,\\Jieji Ren, Wenhai Liu, Weiming Wang, Hao-Shu Fang\\
Shanghai Jiao Tong University}

\let\oldtwocolumn\twocolumn
\renewcommand\twocolumn[1][]{%
    \oldtwocolumn[{#1}{
    \vspace{-10mm}
    \begin{center}
           \includegraphics[width=\textwidth]{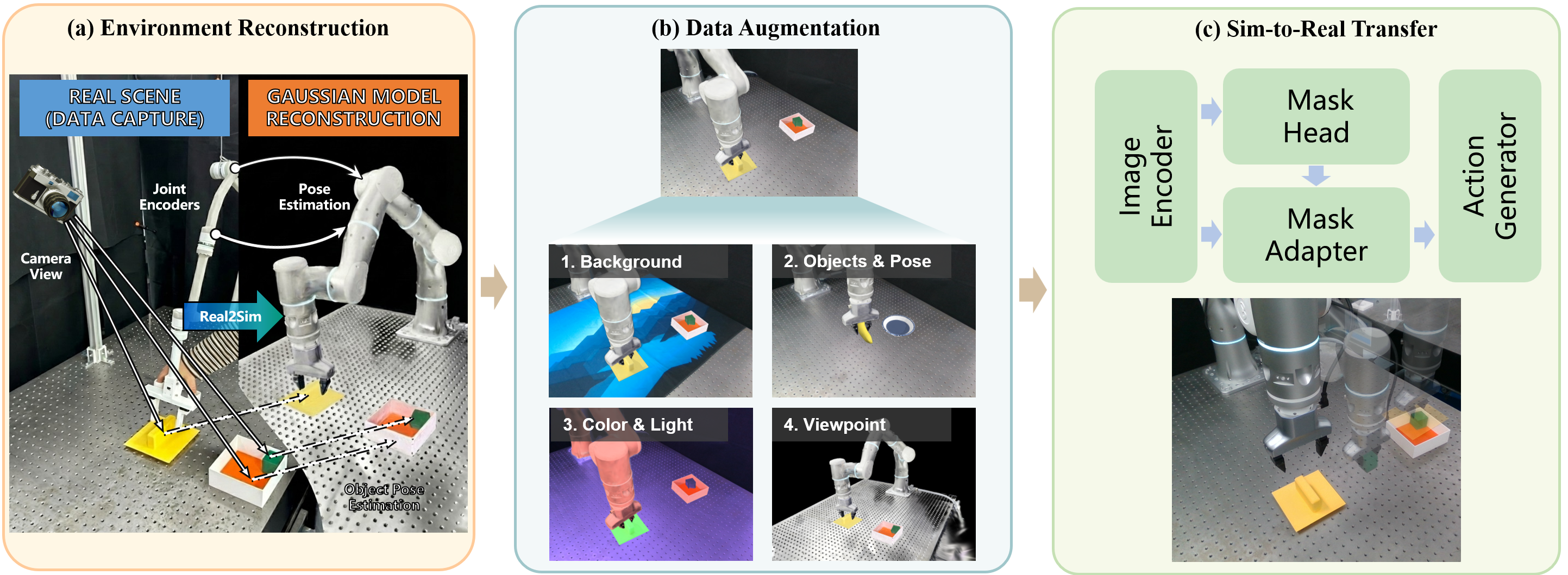}
           \captionof{figure}{Overview of \textbf{ExoGS}. (a) Real-world demonstrations captured by AirExo-3 are reconstructed into editable 3D Gaussian scenes. (b) These assets enable scalable data augmentation across viewpoints and lighting. (c) A semantic Mask Adapter facilitates robust sim-to-real policy transfer for manipulation tasks.}
           \label{fig:teaser}
    \end{center}
    }]
}

\maketitle

\begin{abstract}
Real-to-Sim-to-Real technique is gaining increasing interest for robotic manipulation, as it can generate scalable data in simulation while having narrower sim-to-real gap. However, previous methods mainly focused on environment-level visual real-to-sim transfer, ignoring the transfer of interactions, which could be challenging and inefficient to obtain purely in simulation especially for contact-rich tasks. We propose ExoGS, a robot-free 4D Real-to-Sim-to-Real framework that captures both static environments and dynamic interactions in the real world and transfers them seamlessly to a simulated environment. It provides a new solution for scalable manipulation data collection and policy learning. ExoGS employs a self-designed robot-isomorphic passive exoskeleton AirExo-3 to capture kinematically consistent trajectories with millimeter-level accuracy and synchronized RGB observations during direct human demonstrations. The robot, objects, and environment are reconstructed as editable 3D Gaussian Splatting assets, enabling geometry-consistent replay and large-scale data augmentation. Additionally, a lightweight Mask Adapter injects instance-level semantics into the policy to enhance robustness under visual domain shifts. Real-world experiments demonstrate that ExoGS significantly improves data efficiency and policy generalization compared to teleoperation-based baselines.
Code and hardware files have been released on \url{https://github.com/zaixiabalala/ExoGS}.
\end{abstract}


\section{Introduction}

\IEEEPARstart{I}{mitation} learning has enabled robots to acquire complex manipulation skills from demonstrations. However, its efficacy is heavily based on the scale and quality of the training data. Simulation-based data synthesis offers scalability, but suffers from the well-known sim-to-real gap due to discrepancies in geometric representation, visual appearance, and physical interaction, particularly in visual fidelity for RGB-based policies.

Recent Real-to-Sim-to-Real (R2S2R) pipelines with neural scene representations, such as NeRF and 3D Gaussian Splatting (3DGS)~\cite{mildenhall2021nerf, kerbl20233d}, have shown promise in bridging the visual gap through photorealistic rendering. However, most existing R2S2R pipelines are limited to static scene reconstruction and rely on reinforcement learning to obtain manipulation data. This challenge lies in acquiring physically valid, high-fidelity interaction data without the burden of deploying expensive robotic hardware.

To address this, this paper introduces \textbf{ExoGS}, a low-cost, robot-free R2S2R framework that allows users to capture 4D sequences comprising 3D environments and spatiotemporal robot interactions with objects. Our framework consists of a custom passive exoskeleton with a 3DGS-based data generation pipeline. The core insight is to leverage a robot-isomorphic exoskeleton, AirExo-3, to sensorize and capture human manipulation. The manipulation data including exoskeleton joint configurations and object trajectories are captured by our platform (see Sec.~\ref{sec:collect}) and then digitized into a motion sequence of editable 3DGS assets of the robot and the objects. This hardware-software synergy allows us to decouple the robot, objects, and environment, enabling massive, geometry-consistent data augmentation in simulation. Furthermore, to mitigate the residual domain shift, we introduce a Mask Adapter, a lightweight module that injects instance-level semantic constraints into the policy, guiding attention toward interaction-relevant features.

The main contributions of this work are threefold:
\begin{itemize}
    \item \textbf{A 4D, robot-free R2S2R framework} that reconstructs real-world assets and manipulation sequences into editable 3DGS assets and their dynamics, enabling scalable, embodiment-consistent data generation.
    \item \textbf{AirExo-3}, an open-source, low-cost, accurate, and durable manipulation data collection device.
    \item \textbf{A Mask Adapter} that injects semantic mask information into the policy to steer attention toward interaction-relevant regions, thereby enhancing robustness.
\end{itemize}

\section{Related Works}

\subsection{Sim-to-Real Transfer and 3D Gaussian Splatting}

Due to persistent discrepancies in appearance, geometry, and contact dynamics, policies trained in simulation often fail to generalize to the real world, motivating Real-to-Sim-to-Real (R2S2R) pipelines. R2S2R pipelines reconstruct photorealistic digital twins from real scenes and perform data generation and policy learning in reconstructed environments, leveraging neural radiance fields or 3D Gaussian Splatting for high-fidelity rendering and novel view synthesis~\cite{byravan2023nerf2real,qureshi2024splatsim}. These approaches generally fall into interaction-oriented frameworks that support closed-loop training and rendering-centered pipelines that scale data generation via photorealistic replay~\cite{villasevil2024reconciling,lou2024robo,li2024robogsim,yu2025real2render2real,yuan2025learning}, yet limitations in physical interactability and contact modeling persist.

Compared with NeRF~\cite{mildenhall2021nerf}, 3DGS~\cite{kerbl20233d} offers an explicit, point-based representation with fast optimization and real-time rendering, making it suitable for robotics. With recent extensions incorporating dynamics and physical priors~\cite{wu20244d,xie2024physgaussian}, 3DGS has emerged as an editable, geometry-aware world model for manipulation and navigation~\cite{shorinwasplat,jigraspsplats,deformgs,abou2024physically,lou2024robo}, enabling geometry-consistent edits beyond 2D augmentation~\cite{yang2025novel}. Nevertheless, effectively leveraging 3DGS for robot learning requires tight integration of demonstration acquisition, pose modeling, and interaction-aware replay.

Therefore, a Real-to-Sim-to-Real framework that tightly couples real-world demonstration acquisition, accurate pose modeling, and geometry-consistent 3DGS replay is still needed to support scalable, contact-rich manipulation learning with high visual fidelity, a gap that \textbf{ExoGS} addresses.

\subsection{Demonstration Data Collection}

Demonstration data is fundamental to scalable and generalizable imitation learning in robotic manipulation~\cite{zitkovich2023rt,jang2022bc,zhang2018deep,song2020grasping,bousmalis2023robocat}. Although teleoperation enables embodiment-consistent robot demonstrations, its scalability is fundamentally constrained by hardware cost, deployment overhead, and non-intuitive human-robot interfaces.

To reduce these limitations, an alternative direction focuses on in-the-wild demonstration collection using human-operated devices such as handheld grippers, VR/AR systems, motion-capture gloves, and exoskeletons~\cite{fang2024airexo,fangairexo,chi2024universal,wang2024dexcap,zhaolearning,ben2025homie}. These systems reduce deployment cost and enable more natural demonstrations, but often suffer from limited kinematic fidelity, trajectory accuracy, or environment diversity.

Therefore, existing approaches lack a low-cost, robot-free data collection framework that preserves robot-aligned kinematics and supporting scalable, geometry-consistent replay, motivating the Real-to-Sim-to-Real design of \textbf{ExoGS}.

\section{Method}

ExoGS first employs AirExo-3, a low-cost exoskeleton (detailed in Sec.~\ref{subsec:hardware}), to capture accurate robot manipulation demonstrations without needing a real robot (detailed in Sec.~\ref{subsec:data}~(1)).  Based on these demonstrations, 3D Gaussian Splatting is utilized to replay and augment the demonstrated interactions (detailed in Sec.~\ref{subsec:data}~(2)-(4)). Building upon this real-to-sim dataset, we further propose a novel plug-in module, Mask Adapter, to facilitate more effective sim-to-real transfer of visuomotor policies (detailed in Sec.~\ref{subsec:policy}).

\subsection{Hardware Design: AirExo-3}
\label{subsec:hardware}

AirExo-3 extends prior low-cost exoskeleton-based demonstration systems~\cite{fang2024airexo,fangairexo} by improving kinematic accuracy and ease of deployment. The design of AirExo-3 targets the following objectives:

\begin{itemize}
    \item \textbf{Ease of deployment}: AirExo-3 supports accurate zero-position calibration and straightforward installation without specialized tools, enabled by a simplified mechanical design.
    \item \textbf{High accuracy}: Unlike handheld devices such as UMI~\cite{chi2024universal}, AirExo-3 achieves robust and accurate motion tracking with millimeter-level accuracy via forward kinematics using high-precision rotary encoders, remaining insensitive to visual conditions while enforcing robot-consistent kinematic constraints.
    \item \textbf{Ease of use}: AirExo-3 enables intuitive, contact-rich manipulation through direct gripper interaction, while its lightweight design supports prolonged use with minimal operator fatigue.
\end{itemize}

\vspace{-5mm}

 \begin{figure}[!htbp]
    \centering
    \includegraphics[width=1\columnwidth]{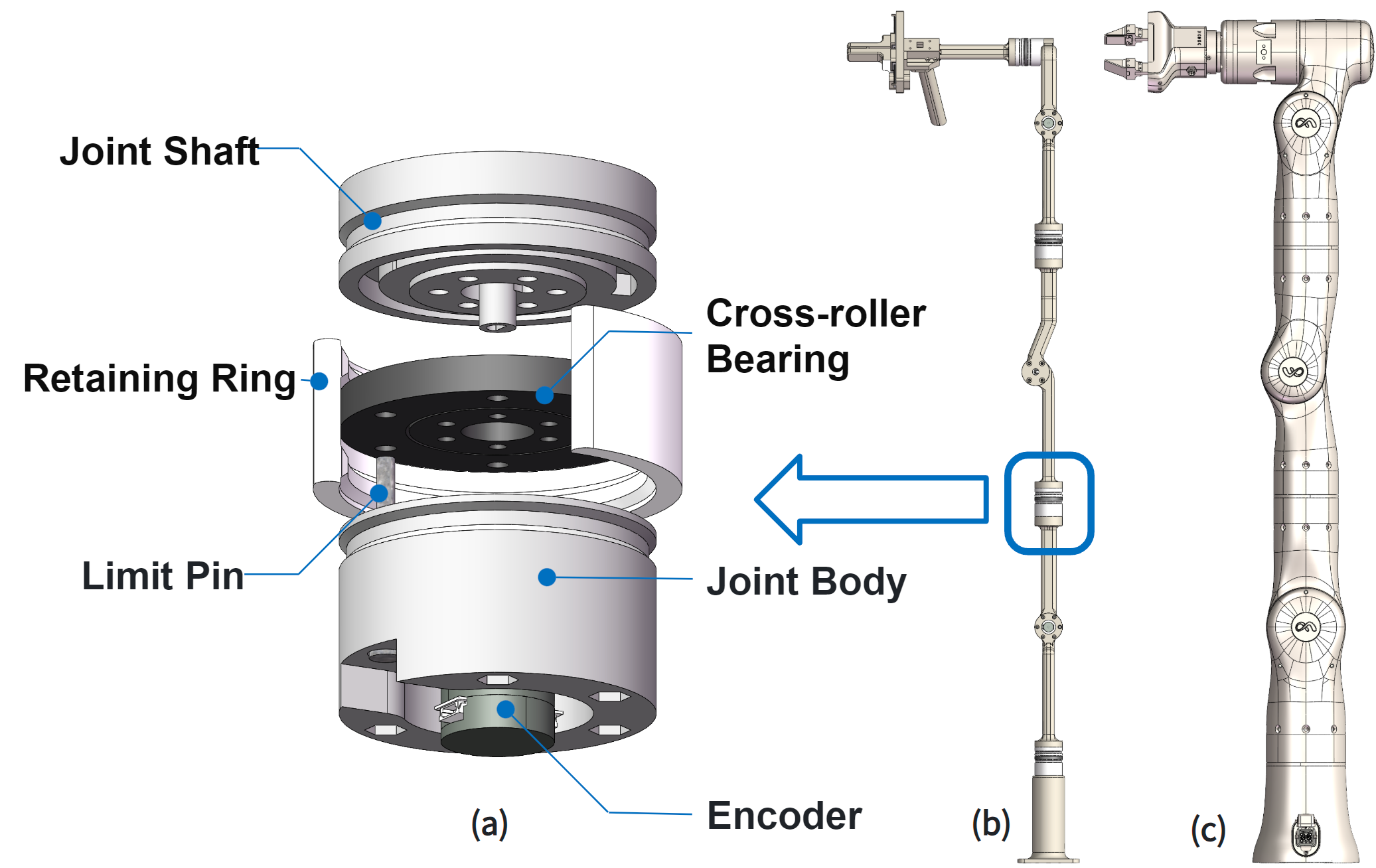}
    \caption{Mechanical structure of the proposed data acquisition device. (a) Structural design of an individual joint module. (b) Overall structure of AirExo-3, consisting of seven articulated joints and a parallel gripper. (c) The target robotic platform used in this work.}
    \label{fig:airexo design}
     \vspace{-1mm}
\end{figure}

 The structure of AirExo-3 is shown in Fig.~\ref{fig:airexo design}. It is a serial link-joint chain geometrically matched to the target robot and equipped with a replaceable handheld gripper, sharing identical kinematic parameters, joint limits, and gripper opening range to ensure workspace consistency. All components are 3D-printed with glass-fiber-reinforced nylon for a lightweight and robust design, with an optional spring-based suspension at Joint 4 to reduce operator effort.

The core component of AirExo-3 is the joint module, which consists of a joint shaft, bearings, and a joint body housing a 12-bit miniature rotary encoder, as shown in Fig.~\ref{fig:airexo design}~(a). Eight encoders are connected via a shared bus, enabling synchronized joint state acquisition at up to approximately 300 Hz. The joint shaft defines the rotational limits, while cross-roller bearings connect the shaft and body, providing high stiffness and rotational accuracy. An established evaluation protocol for exoskeleton-based manipulation systems~\cite{fangairexo} is adopted to assess kinematic accuracy. The results demonstrate that AirExo-3 achieves an average end-effector positioning error of less than 1 mm.

Limit slots are machined into the joint shaft, and a high-strength pin fixed to the joint body engages with the slot to define the joint’s angular limits. When the pin reaches the slot boundary, further rotation is mechanically prevented. The pin hole on the joint body also serves as a reference for zero-position calibration. Calibration is performed by mechanically locking the joint at its zero position using an extended pin and recording the corresponding encoder reading.

Each joint incorporates a groove with a 3D-printed retaining ring to provide tunable passive damping, enabling stable and comfortable operation.

\subsection{4D Data Generation and Augmentation}
\label{subsec:data}

\subsubsection{Manipulation Demonstration Collection with AirExo-3}
\label{sec:collect}

\begin{figure}[!htbp]
    \centering
    \includegraphics[width=1\columnwidth]{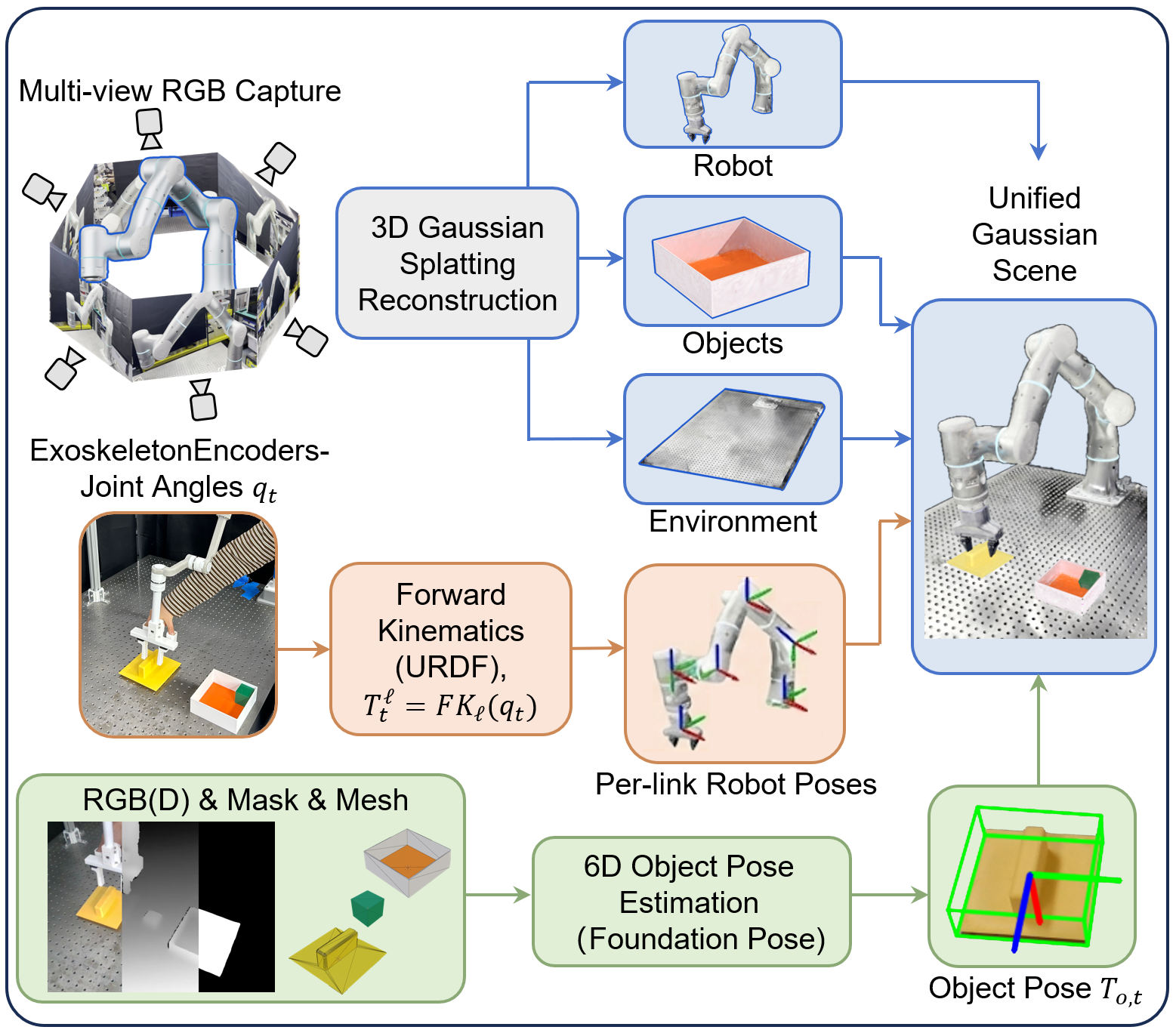}
    \caption{Overview of the pipeline for reconstructing manipulation demonstrations collected with AirExo-3 using 3D Gaussian representations. Camera-based object pose tracking and joint angle encoding from AirExo-3 are combined to generate robot motions, which, together with the reconstructed digital assets, allow the full manipulation process to be faithfully replayed in simulation.}
    \label{fig:gspipeline}
     \vspace{-2mm}
\end{figure}

Let the joint-angle vector of the robotic arm be denoted as $\boldsymbol{q} = [q_1, q_2, \dots, q_n]^\top$, and let the gripper opening be $g \in [0,1]$. Each demonstration trajectory can then be represented as a discrete-time sequence:
\begin{equation}
\tau = \{(\boldsymbol{q}_t, g_t)\}_{t=1}^H,
\end{equation}
where $H$ denotes the number of time steps.

Since AirExo-3 is geometrically and kinematically identical to the target robot, the collected joint states $\boldsymbol{q}_t$ can be directly used for forward kinematics computation, enabling high-fidelity reproduction of manipulation trajectories in both real and simulated environments.

Multiple calibrated Intel RealSense D415 cameras synchronously capture multi-view RGB-D observations $\mathcal{I}=\{I_t^{(k)}\}_{t=1,k=1}^{H,K}$, which are unified in a common world coordinate frame to provide geometric constraints for subsequent 3D reconstruction and pose estimation.

\subsubsection{Digital Asset Generation via Multi-View Reconstruction}
\label{sec:asset}

We adopt 3D Gaussian Splatting (3DGS) to digitize real-world scenes into editable simulation assets. Following the standard formulation~\cite{kerbl20233d}, we represent the scene (including the robot, objects, and environment) as a collection of 3D Gaussians, each parameterized by position, covariance, opacity, and spherical harmonics. We implement a ``capture-reconstruct-assetize" pipeline: multi-view images are first captured and processed via COLMAP~\cite{schoenberger2016sfm} to recover camera poses. These poses initialize the optimization of Gaussian parameters, which is driven by minimizing a weighted L1 and SSIM photometric loss between rendered and captured views. This process yields high-fidelity, decoupled 3D assets for the robotic arm and manipulated objects, enabling independent manipulation and geometry-consistent replay in the simulation environment.

\subsubsection{Object Pose Estimation and Trajectory Processing}
\label{sec:pose}

Multi-view RGB-D sequences are processed with FoundationPose~\cite{foundationCvpr} for object pose tracking. The pose of object $o$ at time $t$ in camera $k$ is denoted as $\mathbf{T}_{o,t}^{(k)} \in SE(3)$. To improve robustness and consistency, we integrate these multi-view estimates by adopting the rotation from the primary camera as the global orientation and averaging the translation vectors across all views. This fusion results in a unified pose sequence $\{\mathbf{T}_{o,t}\}_{t=1}^H$ in the robot base coordinate frame.

To enhance data diversity across task scenarios, we introduce a lightweight pose-processing module, \textit{PoseProcess}, which performs normalization and recomposition of object pose sequences. A \textit{fix} operation constrains the object pose to the robot end-effector frame, enabling rigid attachment during manipulation. By substituting object models, the same pose sequence can be directly transferred to different objects, allowing the generation of diverse task instances without additional data collection.

Robot kinematics are computed using the URDF model of the target robot with the recorded joint-angle sequence $\{\boldsymbol{q}_t\}$. The pose of each link $\ell$ is obtained via forward kinematics:
\begin{equation}
    \mathbf{T}_{\ell,t} = \mathrm{FK}_\ell(\boldsymbol{q}_t), \quad \ell = 1,\dots,L,\; t = 1,\dots,H,
\end{equation}
where $L$ denotes the number of robot links. The resulting robot link poses $\mathbf{T}_{\ell,t}$ are combined with object poses $\mathbf{T}_{o,t}$ in the Gaussian rendering pipeline, enabling multi-view rendering of robot-object scenes for policy training.

\subsubsection{Data Augmentation under Gaussian Rendering}
\label{sec:augmentation}

To improve policy robustness under real-world variations, we employ four data augmentation strategies leveraged by the editable 3DGS representation. First, \textbf{camera viewpoint augmentation} renders demonstrations from perturbed extrinsics to simulate camera placement changes. Second, \textbf{color and illumination augmentation} applies random scaling to Gaussian color attributes and global/local brightness, addressing appearance and lighting gaps. Third, \textbf{background augmentation} composites diverse real-world images as background textures behind the geometry-consistent Gaussian foregrounds, encouraging background-invariant policy learning. Finally, \textbf{object pose augmentation} perturbs object poses and scales, or substitutes objects with affordance-compatible alternatives, enabling trajectory reuse and improving robustness to physical variations.

\subsection{Policy Module: Mask Adapter}
\label{subsec:policy}

\begin{figure*}[t]
    \centering
    \includegraphics[width=1\textwidth]{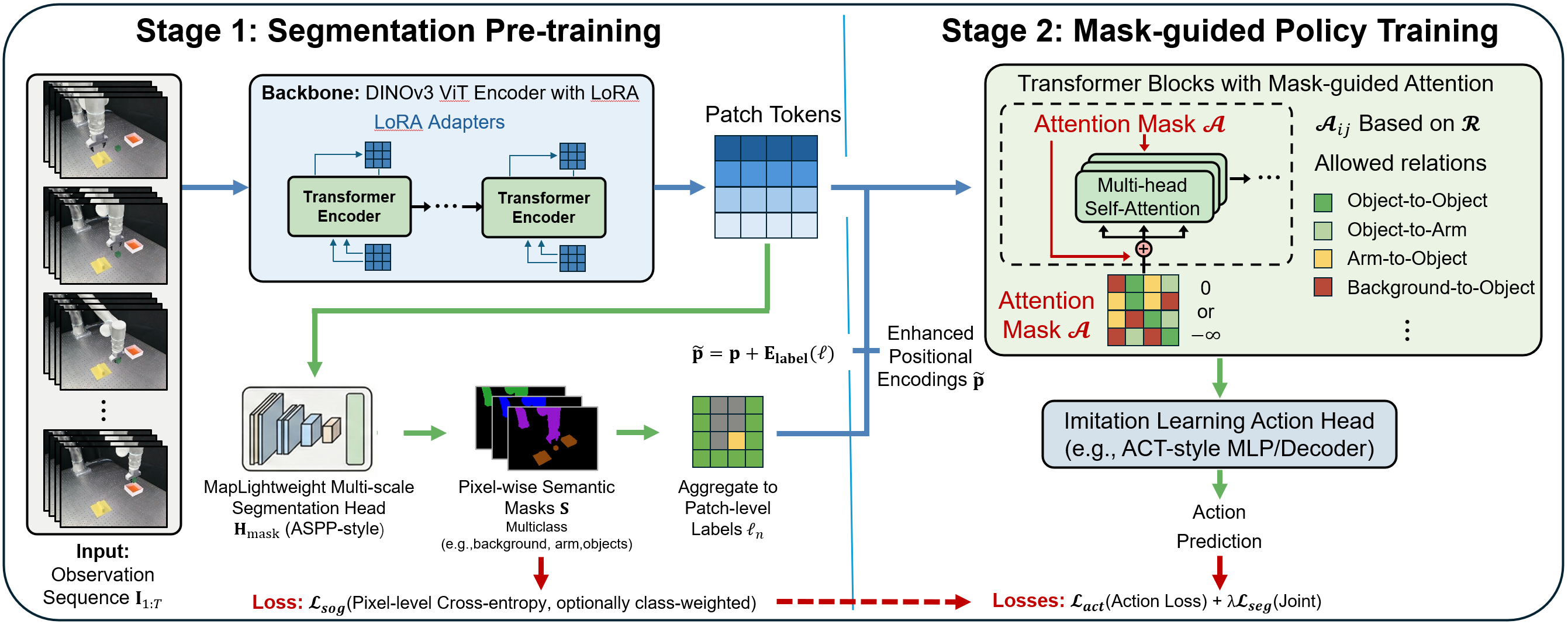}
    \caption{Overview of the proposed Mask Adapter. The module is trained in two stages. Stage 1 performs semantic segmentation pretraining using pixel-level supervision generated by the 3D Gaussian Splatting pipeline, yielding stable patch-level semantic labels for the background, robotic arm, and manipulated objects. Stage 2 incorporates these semantic cues into a ViT-based imitation learning policy via enhanced positional encodings and mask-guided attention, encouraging interaction-relevant token communication to improve robustness and cross-scene generalization under visual domain shifts.}
    \label{fig:maskpipeline}
    \vspace{-5mm}
\end{figure*}

\subsubsection{Introduction} 
\label{sec:actset}

Although in-the-wild demonstrations can be converted into pseudo-robot data, domain gaps in viewpoint, background, and lighting still hinder the generalization of 2D visuomotor policies. Most imitation learning methods lack explicit interaction-centric inductive biases, making them susceptible to spurious background correlations under distribution shifts.

Leveraging the explicit instance-level representation of 3D Gaussian Splatting (3DGS), our pipeline generates pixel-wise semantic masks and supports geometry-consistent data augmentation via multi-view rendering, appearance perturbation, and background replacement, effectively narrowing the sim-to-real gap. Building on an enhanced ACT~\cite{zhaolearning} backbone with a DINOv3 ViT encoder~\cite{simeoni2025dinov3} and LoRA fine-tuning~\cite{hu2022lora}, we introduce Mask Adapter, a lightweight module that injects patch-level semantic cues into ViT-based policies to guide attention toward interaction-relevant regions and improve robustness and cross-scene generalization. The full pipeline is shown in Fig.~\ref{fig:maskpipeline}

Given an observation image sequence $I_{1:T}$, a ViT encoder produces patch tokens and base positional encodings:
$$
\mathbf{x} = E_{\text{vit}}(I_{1:T}) \in \mathbb{R}^{B \times (T N) \times D}, \qquad
\mathbf{p} \in \mathbb{R}^{B \times (T N) \times D},
$$
where $N$ is the number of patches per frame and $D$ is the hidden dimension. Mask Adapter augments this pipeline with a mask head and label-driven attention constraints, trained in two stages.

\subsubsection{Mask Head}
\label{sec:maskhead}

When trained only with action supervision, 2D policies receive sparse and highly task-specific gradients, making it difficult to learn stable semantic structures (e.g., background vs. arm vs. objects). Under domain shifts, this often amplifies attention drift. Therefore, in Stage 1 we fine-tune the visual encoder and a segmentation head with pixel-level supervision to obtain transferable semantics and to provide patch-level labels for Stage 2. 
We adopt a lightweight multi-scale segmentation head $H_{\text{mask}}$, following ASPP~\cite{aspp} style. After reshaping tokens into a feature map $\mathbf{F} \in \mathbb{R}^{B \times D \times h \times w}$, the head predicts pixel logits:
$$
\mathbf{S} = H_{\text{mask}}(\mathbf{F}) \in \mathbb{R}^{B \times C \times H \times W}.
$$
To align with the token sequence, we aggregate pixel predictions into patch-level labels $\ell \in \{0,\dots,C-1\}^{T N}$. Let $\Omega_n$ be the set of pixels belonging to patch $n$; we compute:
$$
\ell_n = \arg\max_{c \in \{0,\dots,C-1\}} \; \frac{1}{|\Omega_n|} \sum_{u \in \Omega_n} \operatorname{softmax}(\mathbf{S}_u)_c.
$$

\subsubsection{Mask-guided Token Modeling}
\label{sec:maskmodel}

Standard Transformers in imitation learning policies typically treat all patches uniformly and provide no structural priors on which tokens should interact. As a result, under occlusions or background changes, the model may aggregate irrelevant context and fail to generalize. In Stage 2, we use patch labels $\ell$ to enhance positional encodings and impose label-driven interaction constraints, encouraging semantic-consistent and interaction-relevant token communication. We add a learnable label embedding to the base positional encoding:
$$
\tilde{\mathbf{p}} = \mathbf{p} + \mathbf{E}_{\text{label}}(\ell).
$$
We define a label relation set $\mathcal{R}$ and construct an additive attention mask $\mathbf{A}$:
$$
\mathbf{A}_{ij} =
\begin{cases}
0, & (\ell_i, \ell_j) \in \mathcal{R} \\
-\infty, & (\ell_i, \ell_j) \notin \mathcal{R}
\end{cases},
$$
which is added to attention logits.

\subsubsection{Training Objective}
\label{sec:maskloss}

We train segmentation with (optionally class-weighted) pixel-level cross-entropy:
$$
\mathcal{L}_{\text{seg}} = -\frac{1}{|\Omega|}\sum_{u \in \Omega} w_{y_u}\log \operatorname{softmax}(\mathbf{S}_u)_{y_u}.
$$
Stage 2 optimizes the original policy action loss $\mathcal{L}_{\text{act}}$ jointly with segmentation to stabilize semantic alignment:
$$
\mathcal{L}_{\text{stage2}} = \mathcal{L}_{\text{act}} + \lambda \, \mathcal{L}_{\text{seg}}.
$$
When ground-truth masks are unavailable, one can optimize $\mathcal{L}_{\text{act}}$ only while still using Stage-1 predictions to provide $\ell$ for positional enhancement and attention constraints. Since Mask Adapter only requires ViT tokens, positional encodings, and an attention interface, it can be integrated into most ViT-based 2D imitation learning policies with minimal architectural changes.

\section{Experiments}
\label{sec:setup}

\begin{figure*}[t]
    \centering
    \includegraphics[width=\textwidth]{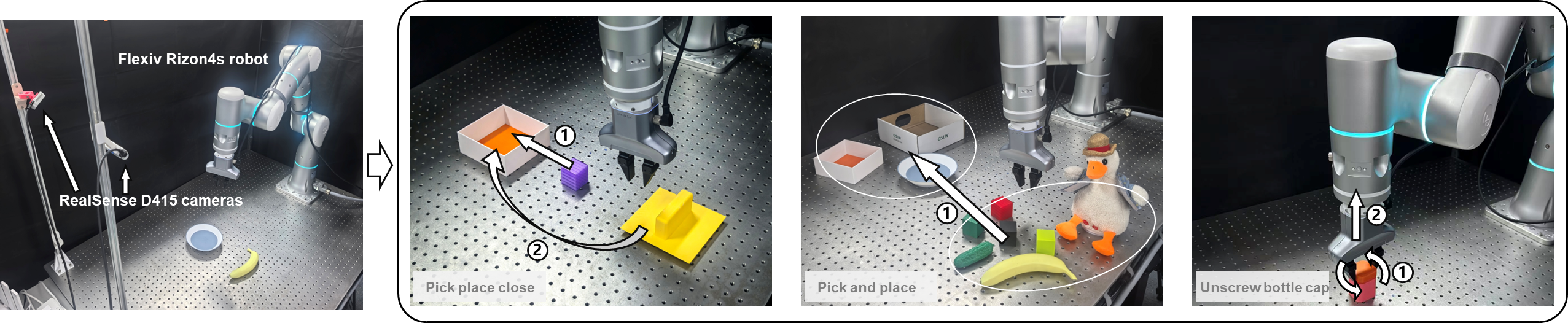}
    \caption{Real-world experimental setup and task illustration. A Flexiv Rizon 4s robotic arm is used together with two eye-on-base Intel RealSense D415 cameras. During standard experiments and teleoperation data collection, only the left camera is used, while the other camera is reserved exclusively for evaluating camera viewpoint variations. Three manipulation tasks are designed in this work, whose detailed descriptions are provided in Sec.~\ref{sec:setup}.}
    \label{fig:task}
    \vspace{-5mm}
\end{figure*}

We conduct multiple real-world experiments to validate the efficiency and effectiveness of our data generation pipeline. The detailed experimental scenarios and task settings are illustrated in Fig.~\ref{fig:task}.

We design three manipulation tasks: \textit{Pick and Place}, \textit{Pick Place Close}, and \textit{Unscrew Bottle Cap}. In \textit{Pick and Place}, the robot picks a target object and places it into a container. In \textit{Pick Place Close}, the robot additionally closes the container lid after placement. In \textit{Unscrew Bottle Cap}, the robot unscrews and removes the cap from a jar fixed to the tabletop, involving contact-rich interactions.

Experiments are conducted on a Flexiv Rizon 4s robotic arm with a Xense Aurora Lite gripper under position control, using an overhead RealSense D415 camera for RGB observations. For each task, policies trained on data generated by our pipeline are compared against those trained on teleoperation data.

For the \textit{Pick and Place} task, objects are randomly placed within a $40\,\text{cm} \times 50\,\text{cm}$ tabletop workspace. Multiple target objects (e.g., cube, fruits, and plush toy) and containers of different materials are considered. Only the cube-box case is physically collected, while all other object-container combinations are synthesized by replacing 3D Gaussian assets in simulation. The \textit{Pick Place Close} task follows the same workspace configuration but requires closing the container lid after placement. In the \textit{Unscrew Bottle Cap} task, the jar is fixed on the table, and the robot unscrews and removes the cap.

Through these experiments, we evaluate ExoGS pipeline efficiency, policy performance, and robustness gains enabled by the proposed augmentation strategies. Across all tasks, we collect 60 raw demonstrations per task using AirExo-3 within our pipeline, and an additional 60 demonstrations via teleoperation for comparison. Data collection time and success rate are reported as evaluation metrics. For policy evaluation, each task is executed over 25 trials, with success rates as the primary performance measure.

\subsection{Evaluation of Demonstration Data Collection Efficiency}

To evaluate data collection efficiency, we recruited 10 volunteers without robotics background, each receiving approximately 10 minutes of training. During data collection, failed attempts were discarded, and only successful demonstrations were used to compute metrics. Each participant recorded six valid demonstrations per task. We report the success ratio and the average time per successful demonstration as evaluation metrics.

 \begin{figure}[htbp]
    \centering
    \includegraphics[width=1\columnwidth]{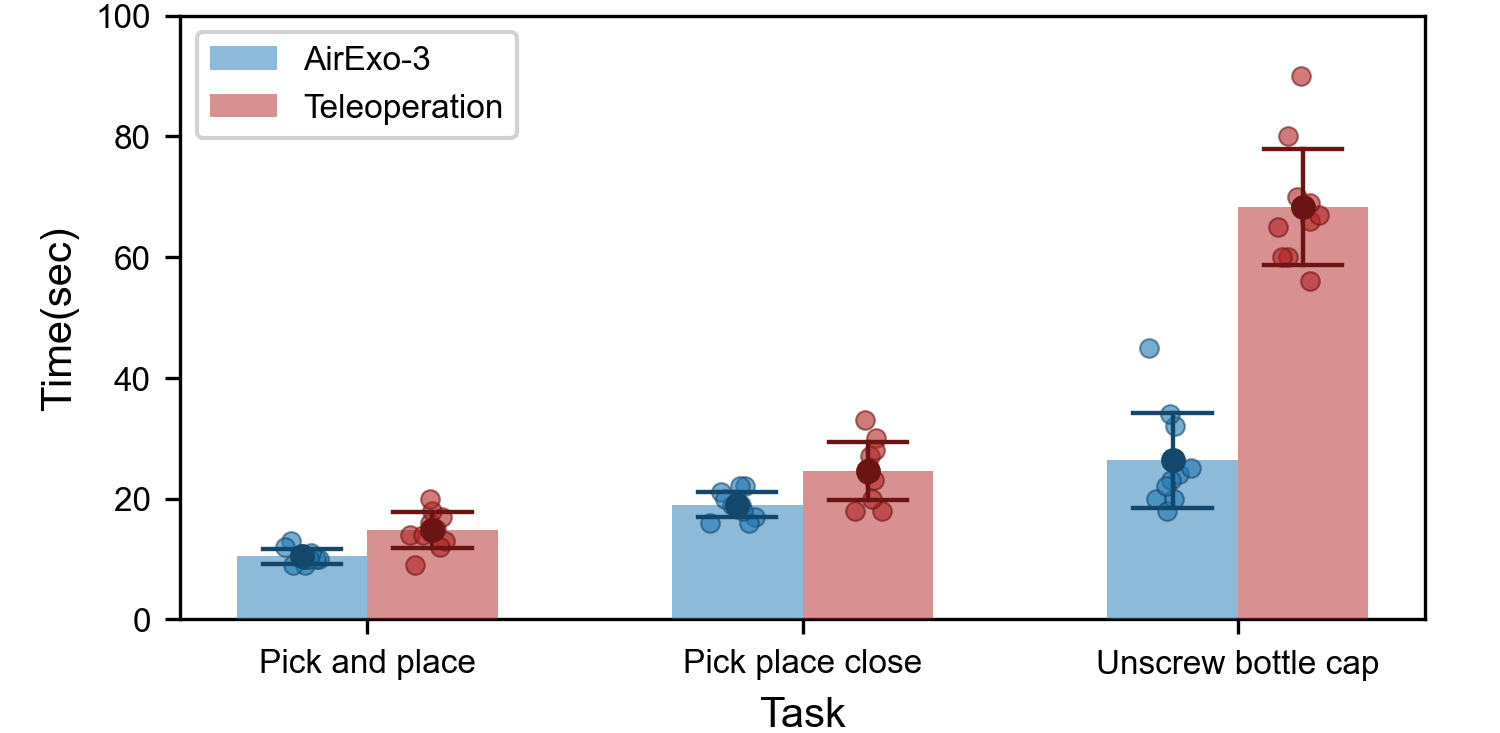}
    \caption{Task completion time comparison between AirExo-3 and teleoperation. Bars show the average over all volunteers, and colored dots denote individual averages, computed using successful trials only.}
    \label{fig:timeuse}
    \vspace{-3mm}
\end{figure}

Fig.~\ref{fig:timeuse} compares average data collection time across manipulation tasks using two acquisition methods. Overall, AirExo-3 achieves faster data collection than teleoperation, with comparable performance on simple tasks and increasingly larger advantages as task complexity grows. This benefit arises from more natural manipulation feedback and lower operational burden, and is further reflected by reduced inter-subject variance, indicating improved consistency across users.

\vspace{-3mm}

\begin{table}[htbp]
\centering
\caption{Task Success Ratio Comparison: AirExo-3 vs. Teleoperation}
\label{tab:task_comparison}
\begin{tabular}{lcc}
\toprule
\textbf{Task} & \textbf{AirExo-3} & \textbf{Teleoperation} \\
\midrule
Pick and place & 100\% & 92.3\% \\
Pick place close & 100\% & 83\% \\
Unscrew bottle cap & 87\% & 17\% \\
\bottomrule
\end{tabular}
\vspace{-3mm}
\end{table}

Table~\ref{tab:task_comparison} reports task success ratios for the two data collection methods. Consistent with the timing results, AirExo-3 achieves higher success ratios than teleoperation across all tasks, with the performance gap increasing as task difficulty grows. The advantage is most pronounced in the bottle-cap unscrewing task, where teleoperation requires many failed attempts and often causes unintended collisions and object damage. In contrast, AirExo-3 yields more stable executions and higher success ratios, demonstrating its effectiveness for collecting reliable demonstrations in contact-rich manipulation tasks.

\subsection{Policy Performance without Data Augmentation}

We evaluate policies trained on data generated by our pipeline against those trained on teleoperation-collected data to assess whether the generated data can match the effectiveness of real-world demonstrations. For the \textit{Pick and Place} task, we further expand the dataset by replacing the manipulated object with different digital assets, yielding a dataset ten times larger than the original. In the \textit{Pick and Place (New Object)} setting, the objects are exclusively drawn from this augmented set and have never appeared during physical data collection, with demonstrations generated solely through trajectory transfer via digital asset replacement. All experiments in this setting are conducted using the modified ACT 
policy described in Sec.~\ref{sec:actset}.

\vspace{-5mm}

\begin{table}[htbp]
\centering
\caption{Success Rate Comparison Between ExoGS and Teleoperation}
\label{tab:sr_compare_vertical_grouped}
\renewcommand{\arraystretch}{1.2}
\small
\begin{tabular}{lcc}
\toprule
\textbf{Task} 
& \textbf{ExoGS} & \textbf{Teleop} \\
\midrule
Pick and place             & 50\% & 72\% \\
Pick place close           & 48\%& 64\% \\
Unscrew bottle cap         & 24\% & 8\% \\
Pick and place (New Object) & 76\%& 0\% \\
\bottomrule
\end{tabular}
\end{table}

\vspace{-3mm}

For most manipulation tasks, policies trained on the original generated demonstrations underperform those trained on teleoperation data, primarily due to the remaining visual gap between rendered and real-world observations. In the \textit{Unscrew Bottle Cap} task, this trend is reversed, likely because teleoperation is particularly difficult, leading to noisy and inconsistent trajectories that hinder effective policy learning.

Moreover, fully synthetic derived demonstrations significantly benefit policy training. By enabling low-cost dataset expansion, these demonstrations allow policies to achieve performance comparable to those trained on real-world data.

\subsection{Policy Performance with Data Augmentation}

Despite lower visual fidelity, the generated data supports flexible augmentation across viewpoints, appearance, backgrounds, and object pose, enabling systematic evaluation of their effects on policy generalization.

 \begin{figure}[htbp]
    \centering
    \includegraphics[width=\columnwidth]{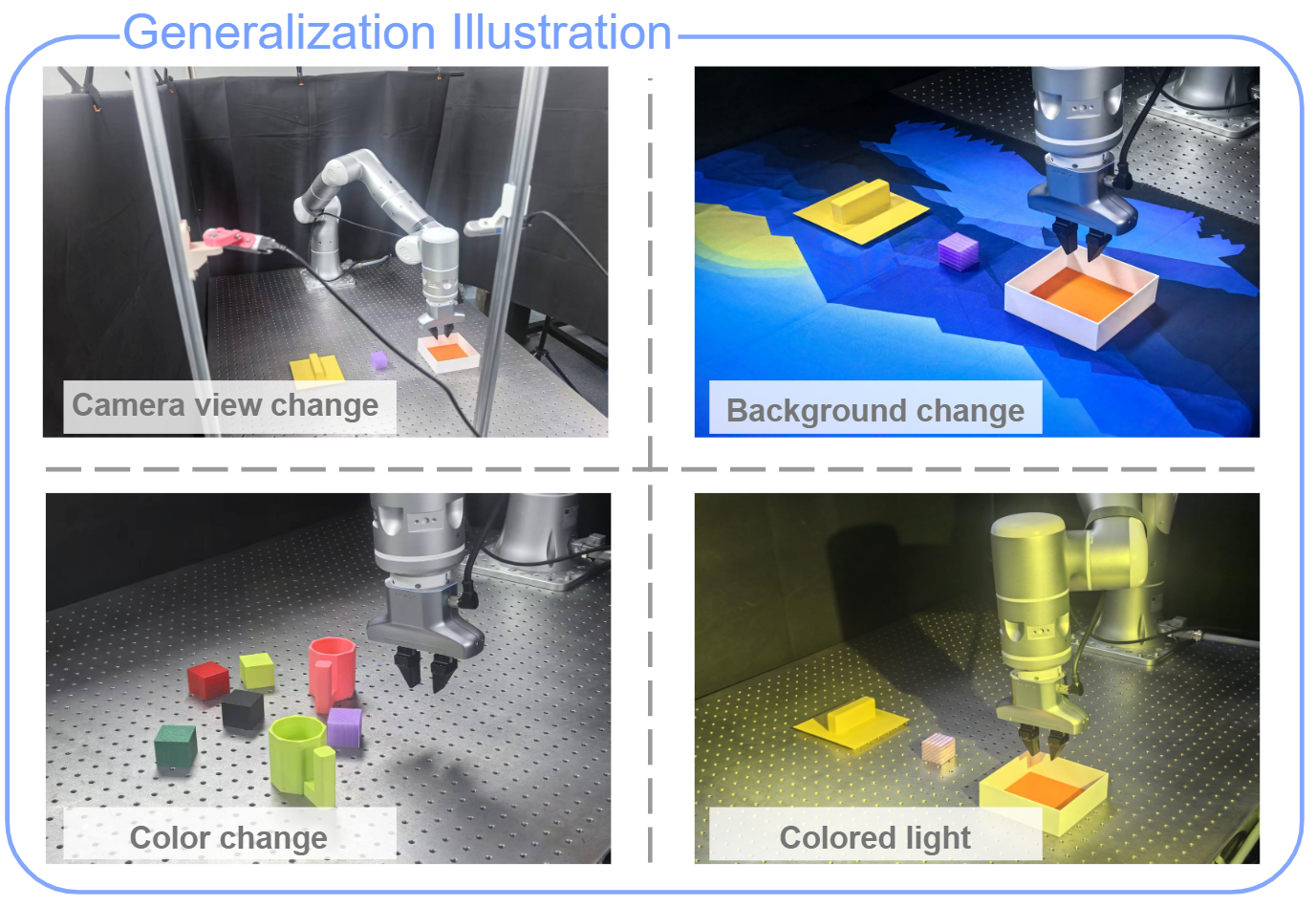}
    \caption{Illustration of the generalization evaluation scenarios. Camera viewpoints, object colors, background appearances, and lighting conditions are varied to assess the generalization capability of the learned policies across different environments.}
    \label{fig:gen}
\end{figure}

Using these four augmentation strategies, we obtain a dataset that is twenty times the size of the original one for policy training. We then introduce variations in both the environment and the manipulated objects to assess model performance under generalization settings. The generalization evaluation scenarios are illustrated in Fig.~\ref{fig:gen}.

\begin{figure*}[t]
    \centering
    \includegraphics[width=0.85\textwidth]{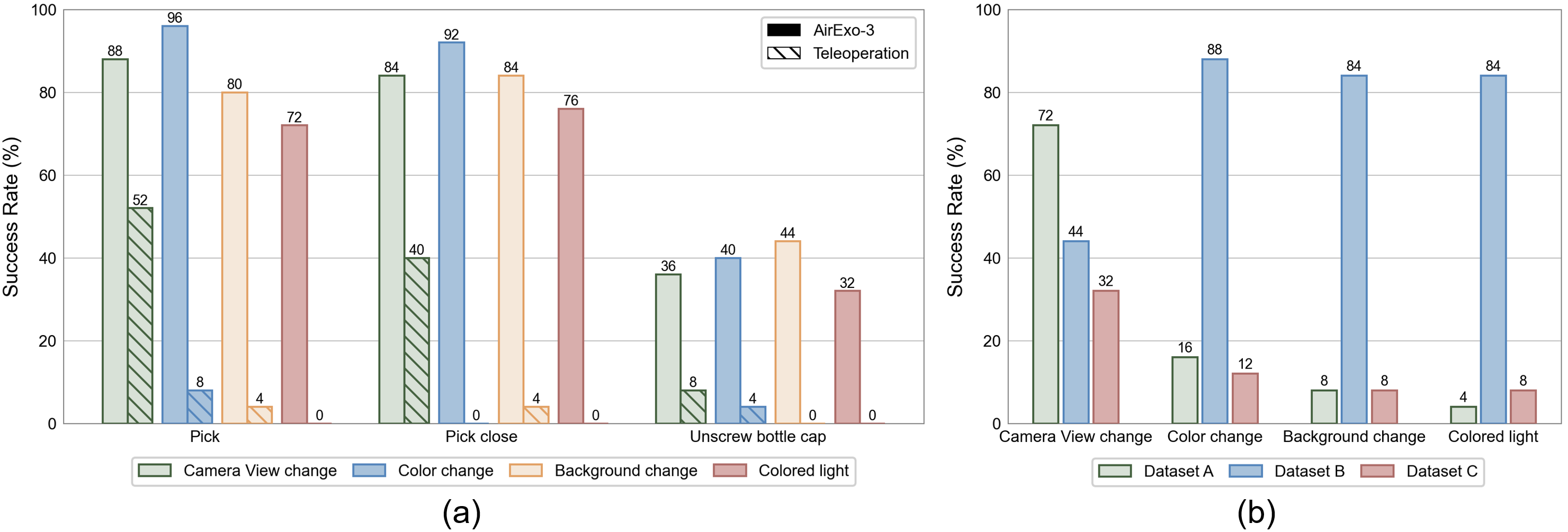}
    \caption{Effect of data augmentation on policy generalization. (a) Success rates under various visual perturbations for policies trained on teleoperation and augmented data. (b) Ablation study using three augmented datasets (A, B, and C) to evaluate the impact of different augmentation strategies on generalization performance.}
    \label{fig:suc_aug}
    \vspace{-5mm}
\end{figure*}

As shown in Fig.~\ref{fig:suc_aug}~(a), data augmentation substantially improves policy generalization. Policies trained with augmented data consistently outperform those trained on non-augmented synthetic data and even real-world data, especially under variations in object color, background, and lighting. In these challenging settings, policies trained solely on real-world data often fail completely, highlighting the effectiveness of the proposed augmentation strategies.

In contrast, the improvement is limited for the \textit{Unscrew Bottle Cap} task, where failures are dominated by strong kinematic constraints from the threaded coupling, such as slippage and jamming, rather than visual perception. These issues mainly arise from suboptimal demonstration quality due to operator proficiency, inherently limiting the benefits of data augmentation.

\subsection{Ablation Study of Data Augmentation Methods}

To evaluate the individual contributions of different data augmentation strategies to policy generalization, we construct three augmented datasets for ablation studies. Since color jitter inherently affects background appearance, it is not considered separately from background replacement. The datasets are defined as follows:

\begin{figure}[htbp]
    \centering
    \includegraphics[width=1\columnwidth]{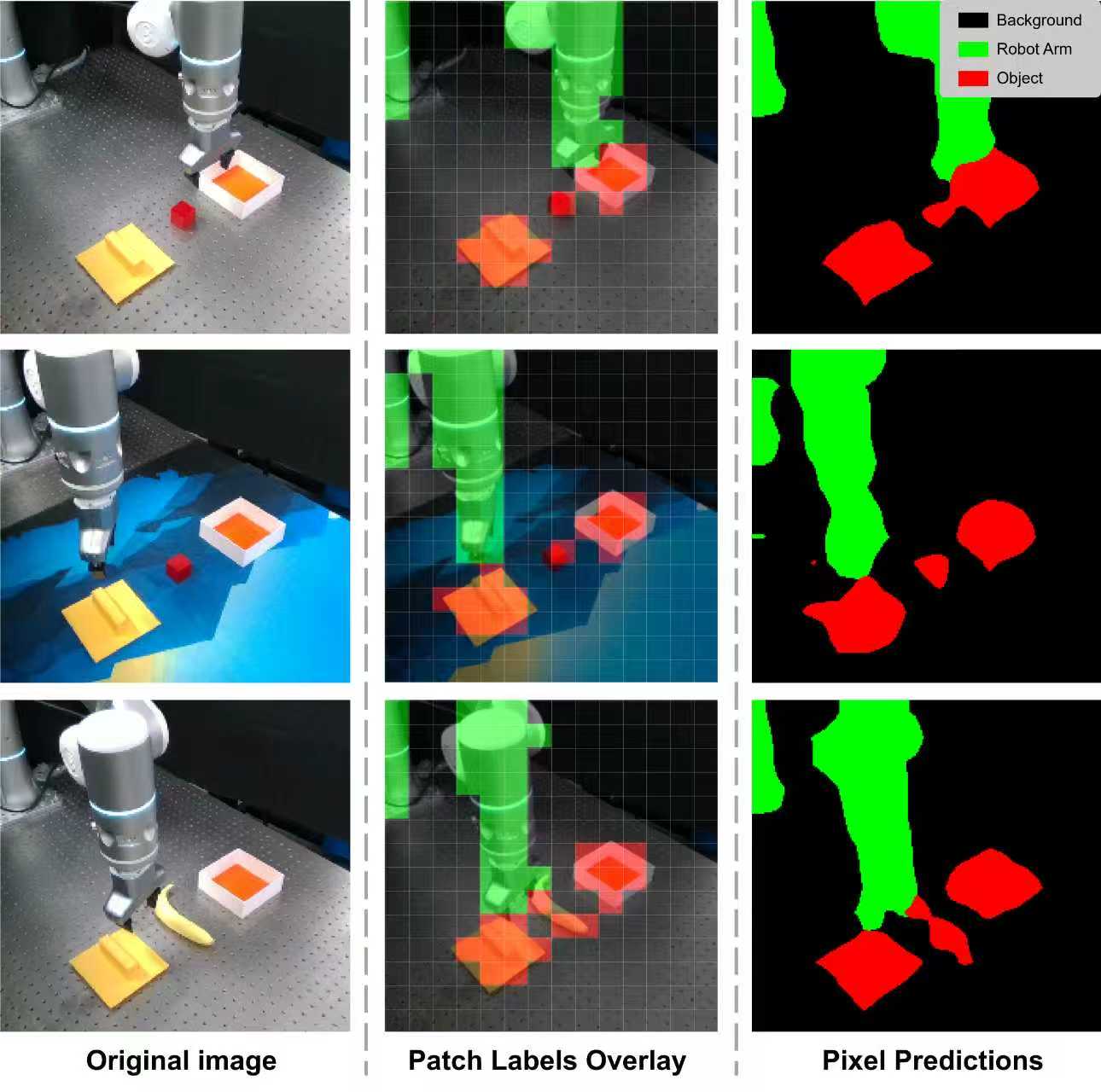}
    \caption{Example segmentation results of the proposed Mask Adapter. The model is trained using only containers and green blocks, yet achieves robust segmentation when tested on novel objects and backgrounds, demonstrating strong generalization capability.}
    \label{fig:maskfig}
    \vspace{-5mm}
\end{figure}

\begin{itemize}
\item \textbf{Dataset A (Viewpoint)}: Each demonstration is rendered from ten novel camera viewpoints using 3DGS, resulting in a dataset ten times the size of the original one.
\item \textbf{Dataset B (Appearance)}: Background textures and lighting conditions are randomized, and large-range color jitter is applied to all Gaussian models, producing a dataset ten times the size of the original one.
\item \textbf{Dataset C (Object Pose)}: Additional demonstrations are generated by applying random perturbations to the poses of the manipulated objects, yielding a dataset ten times the size of the original one.
\end{itemize}

Since the performance bottleneck of the \textit{Unscrew Bottle Cap} task mainly arises from physical constraints rather than visual representation, we conduct the ablation study on the \textit{Pick Place Close} task.

As shown in Fig.~\ref{fig:suc_aug}~(b), the generalization improvements from data augmentation largely align with intuition. Viewpoint variation and color jitter most effectively expand the training domain and thus contribute the largest performance gains, with color jitter yielding the strongest overall improvement due to persistent color and illumination discrepancies between synthetic and real data.

In contrast, pose augmentation provides only marginal benefits, as object poses are already sufficiently diverse during data collection and pose perturbations do not address the primary visual domain gap, resulting in limited performance gains.

\subsection{Effect of Mask Adapter on Policy Performance}

Mask Adapter is a lightweight module designed to enhance policy generalization by guiding attention toward interaction-centric features and facilitating the modeling of interaction-relevant relationships, thereby mitigating the sim-to-real gap. While sharing a similar goal with data augmentation, conventional augmentation relies on substantially enlarged synthetic datasets, incurring notable storage and computational overhead. In contrast, Mask Adapter operates directly on the original data and remains fully compatible with augmented data, offering a more efficient alternative.

We train ACT policies equipped with Mask Adapter using non-augmented demonstrations generated by our pipeline and evaluate their generalization performance under varying environmental and object conditions.
 \vspace{-3mm}

\begin{figure}[htbp]
    \centering
    \includegraphics[width=1\columnwidth]{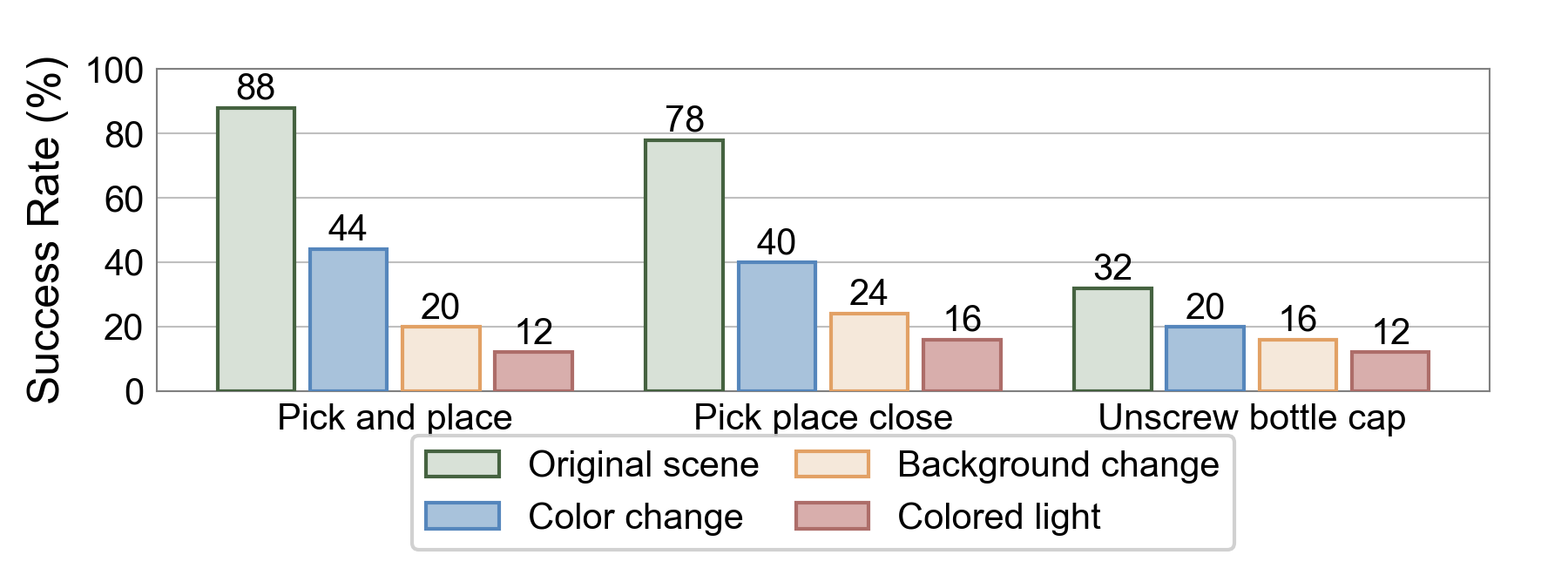}
    \caption{Impact of the Mask Adapter on policy generalization. The adapter effectively narrows the sim-to-real gap, enabling policies to outperform teleoperation baselines in standard and color-varied scenarios, while performance remains sensitive to severe background and lighting perturbations.}
\label{fig:masktest}
\vspace{-1mm}
\end{figure}

Fig.~\ref{fig:maskfig} illustrates segmentation results produced by the Mask Adapter during evaluation. The adapter enables fully automatic and accurate segmentation of the target objects, the background, and the robot, and is able to generalize to objects that do not appear in the training dataset.

As shown in Fig.~\ref{fig:masktest}, the Mask Adapter is highly effective at mitigating the sim-to-real gap introduced by training on synthetic data. Benefiting from the higher-quality motion trajectories collected using AirExo-3, the resulting policies even outperform those trained on teleoperation data. In addition, the policies exhibit a certain degree of robustness to object color variations. However, under more severe conditions, such as drastic background changes and colored lighting that adversely affect the segmentation model, performance still degrades noticeably. Considering that the Mask Adapter is designed as a lightweight module to counter the sim-to-real gap, its performance is fully satisfactory.

\section{Conclusion}

In this work, we presented ExoGS, a scalable, robot-free Real-to-Sim-to-Real framework for manipulation data collection. By integrating a low-cost (approx. \$400) passive exoskeleton with a 3D Gaussian Splatting pipeline, we enable the efficient acquisition of kinematically consistent demonstrations and their digitization into editable simulation assets. This approach allows for massive, geometry-aware data augmentation that significantly enhances policy generalization. Furthermore, our proposed Mask Adapter effectively bridges the sim-to-real gap by injecting instance-level semantic constraints into the policy. Extensive real-world experiments demonstrate that ExoGS reduces data collection costs while yielding policies that outperform those trained via traditional teleoperation. 

However, the current framework relies on the rigid-body assumption of 3DGS, which restricts the modeling of deformable objects with variable geometry. Future work will plan to further improve the automation level of the data generation pipeline.

{\footnotesize
\bibliographystyle{IEEEtran}
\bibliography{refs}
}

\flushbottom

\end{document}